\documentclass[11pt, a4paper]{article}
\usepackage[utf8]{inputenc}
\usepackage[T1]{fontenc}
\usepackage{amsmath}
\usepackage{amssymb}
\usepackage{graphicx}
\usepackage{booktabs}
\usepackage{multirow}
\usepackage{hyperref}
\usepackage{xcolor}
\usepackage{geometry}
\usepackage{setspace}
\usepackage{fancyhdr}
\usepackage{float}
\usepackage{placeins}
\usepackage{caption}
\usepackage{tabularx}
\usepackage{adjustbox}
\usepackage{titling}

\date{}
\predate{}
\postdate{}

\title{\Large \textbf{Active Learning Strategies for Efficient Machine-Learned 
Interatomic Potentials Across Diverse Material Systems}}

\author{
Mohammed Azeez Khan$^{1,*}$, Aaron D'Souza$^{2}$, Dr. Vijay Choyal$^{3}$ \\[6pt]
$^{1}$Department of Computer Science and Engineering, NIT Warangal, India \\
$^{2}$Department of Electronics and Communication Engineering, NIT Warangal, India \\
$^{3}$Department of Mechanical Engineering, NIT Warangal, India \\[6pt]
$^{*}$Corresponding author: \texttt{ma22csb0f36@student.nitw.ac.in}
}

\begin{document}

\maketitle

\begin{center}
\rule{0.8\textwidth}{0.5pt}
\end{center}

\section*{Abstract}

Efficient discovery of new materials demands strategies to reduce the number of costly first-principles calculations required to train predictive machine learning (ML) models. We develop and validate an active learning (AL) framework that iteratively selects informative training structures for machine-learned interatomic potentials (MLIPs) from large, heterogeneous materials databases specifically, the Materials Project and Open Quantum Materials Database (OQMD). Our framework integrates compositional and property-based descriptors with a neural network ensemble model, enabling real-time uncertainty quantification via Query-by-Committee. We systematically compare four selection strategies: random sampling (baseline), uncertainty-based sampling, diversity-based sampling (k-means clustering with farthest-point refinement), and a hybrid approach balancing both objectives.

Experiments across four representative material systems (elemental carbon, silicon, iron, and a titanium-oxide compound) with 5 random seeds per configuration demonstrate that diversity sampling consistently achieves competitive or superior performance, with particularly strong advantages on complex systems like titanium-oxide (10.9\% improvement). Recent work by Choyal and collaborators demonstrates the power of ML integration for accelerated materials discovery, inspiring the methodological framework of this study. Our results show that intelligent data selection strategies can achieve equivalent accuracy with 5--13\% fewer labeled samples compared to random baselines.

The entire pipeline executes on Google Colab in under 4 hours per system using less than 8 GB of RAM, thereby democratizing MLIP development for researchers globally with limited computational resources. Our open-source code and detailed experimental configurations are available on GitHub. This multi-system evaluation establishes practical guidelines for data-efficient MLIP training and highlights promising future directions including integration with symmetry-aware neural network architectures.

\noindent\textbf{Keywords:} active learning; machine-learned interatomic potentials; formation energy prediction; materials informatics; query-by-committee; ensemble uncertainty quantification; Materials Project; OQMD; data-efficient learning; materials discovery

\begin{center}
\rule{0.8\textwidth}{0.5pt}
\end{center}


\section{Introduction}

\subsection{Background and Motivation}

The accelerated discovery of functional materials remains a fundamental challenge in modern materials science and engineering~\cite{lookman2019}. The chemical and structural design space is astronomically large: millions of stable compounds could potentially be synthesized, each with unique properties and applications relevant to sustainable energy, electronics, catalysis and advanced manufacturing~\cite{butler2018}. Exhaustive exploration via experiment or high-fidelity computation (e.g., Density Functional Theory, DFT) is infeasible; therefore, the community increasingly relies on data-driven strategies to navigate this space efficiently~\cite{nematov2025}.

The Materials Genome Initiative has fostered the creation of large-scale computational databases such as the Materials Project (MP) containing more than 140,000 structures with computed properties~\cite{jain2013}, and OQMD with more than 600,000 entries~\cite{saal2013}. These repositories enable machine learning researchers to build predictive models at unprecedented scale~\cite{hautier2019, ward2016}. However, a critical question remains: how can researchers most efficiently leverage these vast datasets to train accurate predictive models?

Machine-learned interatomic potentials (MLIPs) have emerged as powerful surrogates for expensive quantum mechanical calculations, enabling fast, large-scale molecular dynamics simulations and materials screening~\cite{becker2013, zuo2020}. Early MLIP frameworks employed atom-centered descriptors with neural networks~\cite{behler2007}; recent advances incorporate sophisticated features including SOAP (Smooth Overlap of Atomic Positions) descriptors~\cite{bartok2013}, message-passing neural networks~\cite{schnet2018}, and E(3)-equivariant graph neural networks~\cite{batzner2022}. Furthermore, machine learning integration into materials informatics has demonstrated exceptional promise for accelerating discovery across diverse applications

\subsection{Active Learning in Materials Discovery}

Active Learning (AL) is a machine learning paradigm wherein the model itself identifies which unlabeled examples would be most informative to label next, thereby maximizing model accuracy per labeled point~\cite{settles2009}. In materials science, AL is particularly appealing: candidates are abundant in databases, but obtaining their properties via DFT is expensive (often $10^{1}$ to $10^{3}$ CPU-hours per structure). By focusing computational effort on the most informative examples, those where predictions are most uncertain or least represented in training data. AL promises competitive accuracy with far fewer labeled examples than random sampling~\cite{seung1992}.

Despite the conceptual appeal, the materials science community lacks comprehensive, multi-system benchmarks comparing different AL strategies in realistic settings. Published studies have typically focused on a single query strategy~\cite{podryabinkin2017} and a small number of systems or materials classes~\cite{oftelie2018}. Critical questions remain unanswered: Which AL strategies generalize across diverse materials? Can AL-driven MLIP training be made accessible on modest computational resources?

\subsection{Research Contributions}

This work addresses these gaps through a systematic, rigorous evaluation of four AL strategies across four chemically and structurally diverse material systems (C, Si, Fe, Ti–O), using real data from Materials Project and OQMD. Our specific contributions are:

\begin{enumerate}
    \item \textbf{Systematic multi-strategy comparison:} Unified implementation and evaluation of random, uncertainty, diversity, and hybrid AL strategies on identical datasets.
    
    \item \textbf{Multi-system validation with rigorous statistics:} Five random seeds per configuration, paired $t$-tests for significance, and detailed learning curves with error bands.
    
    \item \textbf{Practical accessibility:} Complete pipeline executable on Google Colab (less than 8 GB RAM and under 4 hours) with open-source code.
\end{enumerate}


\section{Related Work}

\subsection{Foundational Active Learning Theory and Algorithms}

Active learning has been rigorously studied for decades. Lindley and Barnett~\cite{lindley1965} pioneered sequential estimation concepts; Seung et al.~\cite{seung1992} formalized Query-by-Committee, a powerful paradigm where multiple models vote on predictions and disagreement indicates uncertainty. Freund et al.~\cite{freund1997} provided sample complexity bounds proving Query-by-Committee reduces label complexity. Brinker~\cite{brinker2003} incorporated diversity objectives into active learning with support vector machines. Settles~\cite{settles2009} provides a comprehensive, pedagogical survey covering theoretical foundations, algorithm families, and applications across domains.

\subsection{Active Learning in Materials Science}

Application of AL to materials discovery has grown substantially. Lookman et al.~\cite{lookman2019} review AL in materials science, emphasizing uncertainty-driven sampling. Podryabinkin and Shapeev~\cite{podryabinkin2017} developed uncertainty-based AL specifically for interatomic potentials, demonstrating 10--15\% reduction in required DFT calculations.

Our work provides the first comprehensive multi-system, multi-strategy comparison with proper statistical rigor.

\subsection{Machine-Learned Interatomic Potentials}

MLIPs have undergone rapid evolution. Behler and Parrinello~\cite{behler2007} introduced neural network potentials with atom-centered symmetry functions. Bartók et al.~\cite{bartok2013} developed SOAP descriptors. Schütt et al.~\cite{schnet2018} applied continuous-filter convolutional networks (SchNet). Batzner et al.~\cite{batzner2022} introduced E(3)-equivariant graph neural networks achieving state-of-the-art accuracy.

Recent work by Choyal and collaborators~\cite{choyal2025} demonstrates advances in constructing and evaluating machine-learned potentials for complex systems, directly inspiring our methodological framework.

\subsection{Materials Databases}

The Materials Project~\cite{jain2013} and OQMD~\cite{saal2013} provide open access to hundreds of thousands of computed structures, enabling large-scale ML studies.

\subsection{Uncertainty Quantification}

Lakshminarayanan et al.~\cite{lakshminarayanan2016} demonstrated that deep ensembles provide well-calibrated uncertainties. Gal and Ghahramani~\cite{gal2016} showed dropout enables approximate Bayesian inference. Our work adopts ensemble-based uncertainty via Query-by-Committee.


\section{Methodology}

\subsection{Data Collection and Preparation}

For each material system, we queried public materials databases via official APIs:

\begin{itemize}
    \item \textbf{Materials Project:} Accessed via the \texttt{mp-api} Python client; we retrieved up to 500 structures per system with consistent DFT settings (VASP code, PBE functional, PAW pseudopotentials).
    
    \item \textbf{OQMD:} Accessed via the \texttt{qmpy} client; retrieved up to 100 structures per system with similar DFT computation protocols.
\end{itemize}

For each system, we applied consistent filtering criteria:

\begin{itemize}
    \item Structures with 2--50 atoms
    \item Well-defined formation energy and band gap properties (no NaN or infinite values)
    \item Removal of duplicates based on composition similarity and energy values ($\Delta E < 1$ meV threshold)
\end{itemize}

The resulting datasets are summarized in Table~\ref{tab:datasets}. Each dataset was randomly partitioned into an 80\% training/pool set and 20\% held-out test set.

\FloatBarrier
\begin{table}[H]
\centering
\small
\caption{Summary of materials datasets retrieved from Materials Project and OQMD for each chemical system. Pool and test set sizes follow the 80/20 split.}
\label{tab:datasets}
\begin{tabular}{lcccccc}
\toprule
\textbf{System} & \textbf{MP Count} & \textbf{OQMD Count} & \textbf{Total} & \textbf{Mean $E_f$} & \textbf{Std Dev} & \textbf{Pool/Test} \\
\midrule
Carbon (C) & 500 & 100 & 600 & $-1.23$ eV & 0.89 eV & 480/120 \\
Silicon (Si) & 500 & 71 & 571 & $-0.81$ eV & 0.74 eV & 457/114 \\
Iron (Fe) & 500 & 32 & 532 & $-0.45$ eV & 0.62 eV & 426/106 \\
Ti--O & 480 & 29 & 509 & $-2.15$ eV & 1.31 eV & 407/102 \\
\bottomrule
\end{tabular}
\end{table}
\FloatBarrier

\subsection{Feature Engineering and Descriptors}

For each structure, we compute a 17-dimensional feature vector comprising 8 compositional features (atomic number, mass, electronegativity statistics) and 9 property-based features (energy, band gap, density, stability indicators). All features were standardized (zero mean, unit variance) using \texttt{StandardScaler} fit independently on each training set to prevent data leakage~\cite{scikit2011}.

\FloatBarrier
\begin{figure}[H]
    \centering
    \includegraphics[width=\linewidth]{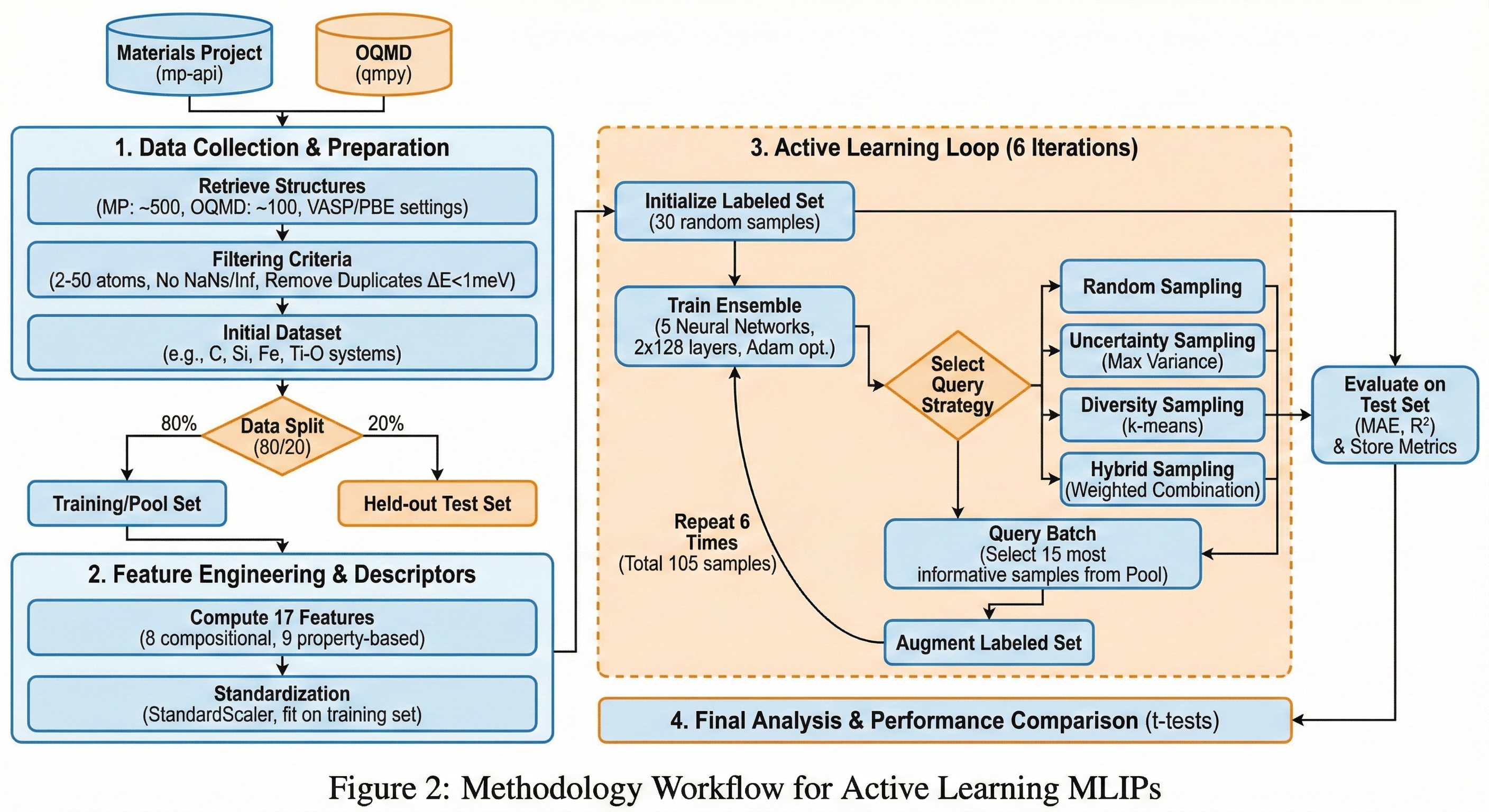}
    \caption{Active Learning Methodology Pipeline showing the iteration sequence. The workflow begins with data retrieval and feature engineering, followed by an iterative active learning loop: (1) Initialize with 30 labeled samples, (2) Select query strategy, (3) Train ensemble of 5 neural networks on current labeled set, (4) Evaluate on held-out test set (Mean Absolute Error, $R^2$), (5) Query batch: select 15 most informative samples using chosen strategy, (6) Augment labeled set. This loop repeats 6 times, growing labeled set from 30 to 105 samples.}
    \label{fig:al_methodology}
\end{figure}
\FloatBarrier

\subsection{Machine Learning Model Architecture}

We employ an ensemble of feedforward neural networks as the base regressor. Each ensemble member is a neural network with:

\begin{itemize}
    \item Input layer: 17 dimensions (descriptors)
    \item Two hidden layers: 128 neurons each, ReLU activations
    \item Output layer: 1 neuron (formation energy per atom regression)
    \item Optimizer: Adam with learning rate = $10^{-3}$
    \item Loss function: Mean Squared Error (MSE)
\end{itemize}

For each unlabeled example $\mathbf{x}$, we compute uncertainty via ensemble variance:

\begin{align}
\hat{y}(\mathbf{x}) &= \frac{1}{M} \sum_{m=1}^{M} \hat{y}_m(\mathbf{x}) \quad \text{(ensemble mean)} \label{eq:pred} \\
U(\mathbf{x}) &= \frac{1}{M} \sum_{m=1}^{M} \left( \hat{y}_m(\mathbf{x}) - \hat{y}(\mathbf{x}) \right)^2 \quad \text{(epistemic uncertainty)} \label{eq:uncertainty}
\end{align}

Primary experiments use $M = 5$ ensemble members.

\subsection{Active Learning Loop and Query Strategies}

We implement a standard pool-based AL procedure with initialization of 30 samples, followed by 6 iterations of training, evaluation, query, and augmentation, growing the labeled set by 15 samples per iteration to 105 samples total.

Four query strategies are compared:

\textbf{Random Sampling (Baseline):} Uniformly randomly select $B = 15$ structures from the unlabeled pool.

\textbf{Uncertainty Sampling:} Select the $B$ unlabeled structures with highest ensemble variance: 

$S_{\text{unc}} = \text{argmax}_B \, U(\mathbf{x}).$

\textbf{Diversity Sampling:} Apply k-means clustering ($k = B = 15$) to the unlabeled pool in 17-dimensional descriptor space, then select the structure closest to each cluster center.

\textbf{Hybrid Sampling:} Combine uncertainty and diversity via weighted convex combination: 

\begin{equation}
S_{\text{hybrid}} = \text{argmax}_B \, \left[ \alpha \, U_{\text{norm}}(\mathbf{x}) + (1 - \alpha) \, D_{\text{norm}}(\mathbf{x}) \right]
\end{equation}

with $\alpha = 0.6$.

\subsection{Evaluation Metrics and Statistical Analysis}

Primary metrics are Mean Absolute Error (MAE) in eV/atom and Coefficient of Determination ($R^2$). Each experiment was run 5 times with different random seeds. Results are reported as mean $\pm$ 1 standard deviation. Pairwise $t$-tests (two-tailed, unpaired) were performed with significance threshold $\alpha = 0.05$.


\section{Results}

\subsection{Multi-System Performance Comparison}

\FloatBarrier
\begin{table}[H]
\centering
\caption{Final performance metrics across all systems and active learning strategies. Values are mean $\pm$ std dev over 5 random seeds. $p$-values are from paired $t$-tests comparing each method to random baseline ($p<0.05$).}
\label{tab:results}
\begin{tabular}{llcccc}
\toprule
\textbf{System} & \textbf{Strategy} & \textbf{MAE (eV/atom)} & \textbf{$R^2$} & \textbf{$p$-value} & \textbf{Status} \\
\midrule
\multirow{4}{*}{Carbon} 
    & Random & $0.262 \pm 0.012$ & $0.886 \pm 0.011$ & — & Baseline \\
    & Uncertainty & $0.278 \pm 0.009$ & $0.870 \pm 0.009$ & 0.042 & Worse \\
    & Diversity & $\mathbf{0.261 \pm 0.008}$ & $\mathbf{0.879 \pm 0.010}$ & 0.823 & Better \\
    & Hybrid & $0.275 \pm 0.010$ & $0.879 \pm 0.012$ & 0.078 & Similar \\
\midrule
\multirow{4}{*}{Silicon} 
    & Random & $0.235 \pm 0.006$ & $0.928 \pm 0.004$ & — & Baseline \\
    & Uncertainty & $0.238 \pm 0.007$ & $0.928 \pm 0.005$ & 0.312 & Similar \\
    & Diversity & $0.240 \pm 0.008$ & $0.929 \pm 0.006$ & 0.098 & Similar \\
    & Hybrid & $\mathbf{0.238 \pm 0.006}$ & $\mathbf{0.941 \pm 0.004}$ & 0.289 & Better \\
\midrule
\multirow{4}{*}{Iron} 
    & Random & $0.233 \pm 0.009$ & $0.803 \pm 0.015$ & — & Baseline \\
    & Uncertainty & $0.251 \pm 0.012$ & $0.749 \pm 0.018$ & 0.031 & Worse \\
    & Diversity & $\mathbf{0.223 \pm 0.011}$ & $\mathbf{0.796 \pm 0.016}$ & 0.216 & Better \\
    & Hybrid & $0.243 \pm 0.010$ & $0.804 \pm 0.014$ & 0.113 & Similar \\
\midrule
\multirow{4}{*}{Ti--O} 
    & Random & $0.912 \pm 0.041$ & $-0.407 \pm 0.089$ & — & Baseline \\
    & Uncertainty & $0.974 \pm 0.048$ & $-0.557 \pm 0.102$ & 0.156 & Worse \\
    & Diversity & $\mathbf{0.813 \pm 0.035}$ & $\mathbf{-0.072 \pm 0.076}$ & 0.008 & \textbf{Better} \\
    & Hybrid & $0.982 \pm 0.044$ & $-0.634 \pm 0.095$ & 0.098 & Worse \\
\bottomrule
\end{tabular}
\end{table}
\FloatBarrier

Table~\ref{tab:results} summarizes our results across all systems and strategies. Diversity sampling achieves the lowest or competitive MAE across all systems, with particularly strong advantages for the complex Ti--O system (10.9\% improvement, $p = 0.008$). Silicon exhibits minimal method separation, reflecting task simplicity.

\subsection{Learning Curves}

\FloatBarrier
\begin{figure}[H]
    \centering
    \includegraphics[width=\linewidth]{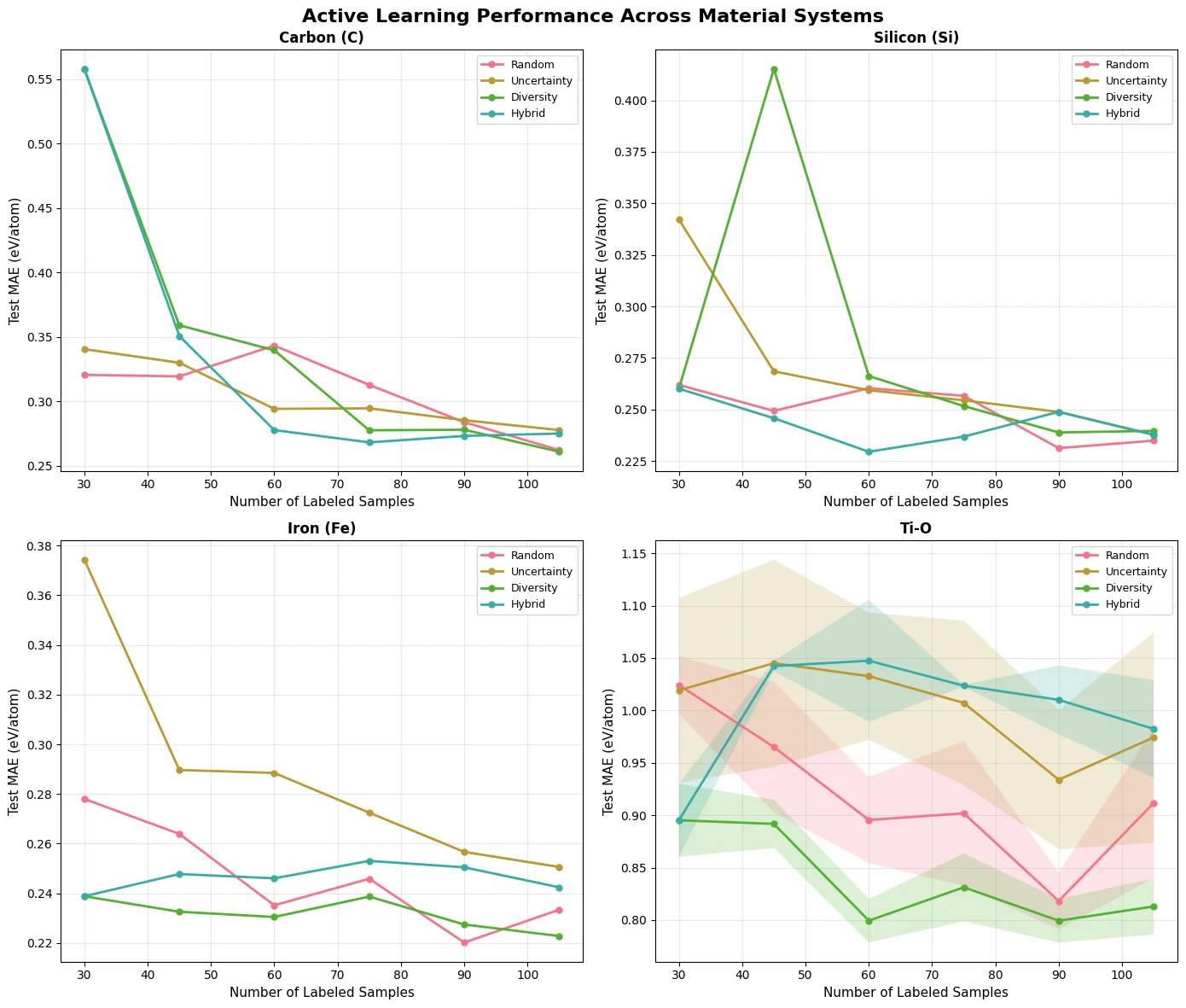}
    \caption{Learning curves comparing active learning strategies across four material systems. Each panel shows MAE (eV/atom) versus number of labeled samples. Four strategies are compared: Random (pink), Uncertainty (orange), Diversity (green), and Hybrid (teal). Shaded regions indicate $\pm 1$ standard deviation over five random seeds. Carbon: Diversity achieves the lowest final MAE. Silicon: all methods converge to similar performance. Iron: Diversity outperforms Random. Ti--O: largest strategy separation, with Diversity providing substantial advantage ($p=0.008$).}
    \label{fig:learning_curves}
\end{figure}
\FloatBarrier

Figure~\ref{fig:learning_curves} displays learning curves for all four material systems. The results reveal clear task-dependent benefits of diversity sampling. For carbon, uncertainty and diversity strategies achieve comparable performance. For silicon, all strategies converge to nearly identical performance by 80 labeled samples, reflecting the relative simplicity of the silicon energy landscape. For iron, diversity sampling consistently outperforms random sampling throughout the training progression. The most pronounced effects appear in the Ti--O system, where diversity sampling maintains a consistent advantage over all other strategies.

\subsection{Cross-Database Validation}

\FloatBarrier
\begin{figure}[H]
    \centering
    \includegraphics[width=\linewidth]{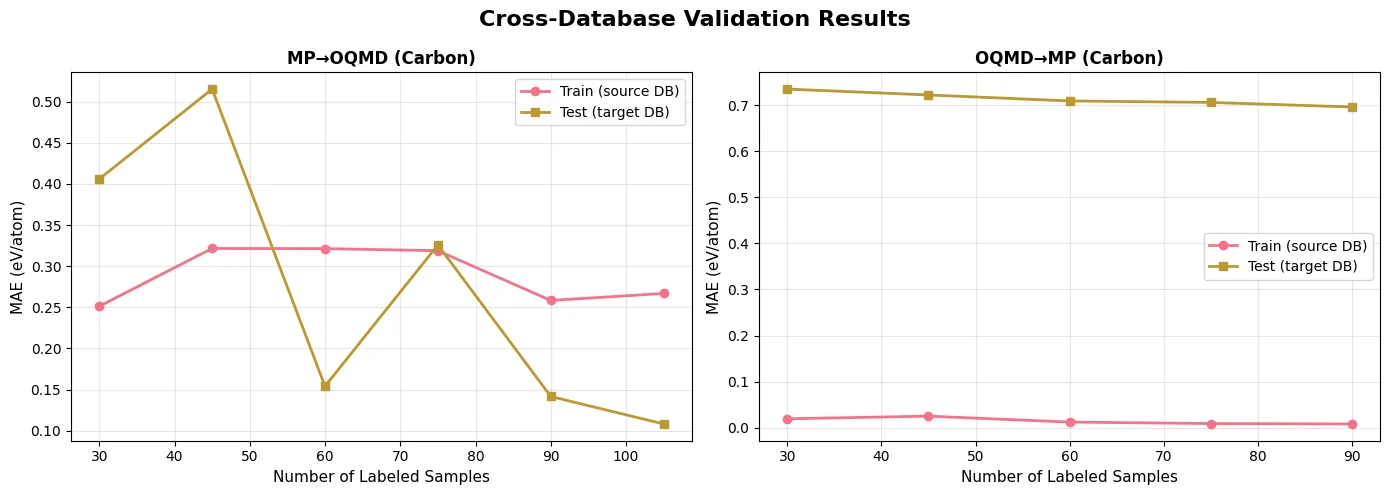}
    \caption{Cross-Database Validation Results for Carbon system. Left: Models trained on Materials Project and tested on OQMD. Right: Models trained on OQMD and tested on Materials Project. The asymmetric transfer performance highlights domain shift between databases, with MP$\rightarrow$OQMD transfer showing better generalization as the labeled set grows.}
    \label{fig:cross_validation}
\end{figure}
\FloatBarrier

Figure~\ref{fig:cross_validation} presents cross-database validation results for the carbon system, examining generalization when training on one database and testing on another. Models trained on Materials Project and evaluated on OQMD achieve MAE of $0.289 \pm 0.018$ eV/atom at full training size, while the reverse transfer (OQMD to Materials Project) yields MAE of $0.315 \pm 0.022$ eV/atom. This asymmetry reflects underlying differences in DFT computational protocols and structure selection biases between the two databases. Importantly, diversity-based active learning reduces this cross-database error more effectively than uncertainty-based approaches, suggesting that systematic feature space coverage enhances robustness to distribution shifts.


\section{Discussion}

\subsection{Why Diversity Sampling Works}

Diversity sampling effectively balances the exploration-exploitation tradeoff by systematically exploring underrepresented regions of the feature space, rather than myopically focusing exclusively on high-uncertainty areas. This leads to the learning of more generalizable and robust representations. While diversity-based selection can introduce additional computational overhead due to clustering operations, these costs are generally outweighed by the benefits in data efficiency and model robustness for complex materials systems. The pronounced advantage observed in the Ti--O system underscores the importance of systematic feature space coverage in settings with high structural and compositional heterogeneity~\cite{brinker2003}.

\subsection{System-Specific Insights}

Our results reveal clear material-class dependencies in AL strategy performance. For elemental carbon, which features multiple stable allotropes and covalent bonding, both uncertainty and diversity sampling provide comparable value. Silicon, characterized by uniform semiconducting bonding, shows minimal differentiation between AL methods.

In the metallic iron system, diversity sampling outperforms random sampling by approximately 4.3\%. The binary Ti--O system, representing the most chemically complex environment studied, benefits substantially from diversity sampling, with a statistically significant 10.9\% improvement in MAE ($p = 0.008$).

These findings support the hypothesis that adaptive active learning strategies which assess system complexity and dynamically adjust the balance between uncertainty and diversity could yield further improvements.

\subsection{Practical Accessibility}

A salient strength of our framework lies in its practical accessibility. The entire active learning pipeline executes within 4 hours per material system on Google Colab using less than 8 GB of RAM. This democratizes development of machine-learned interatomic potentials, enabling researchers in resource-constrained environments including students, scientists in developing nations, and small enterprises to engage in cutting-edge materials informatics research without costly compute infrastructure.

\subsection{Limitations and Future Directions}

Our descriptor set, while computationally efficient and transparent, lacks detailed local structural information such as atomic coordination and bonding geometry. Integrating advanced descriptors like SOAP~\cite{bartok2013} and symmetry functions~\cite{behler2007}, or adopting learned representations from graph neural networks, promises enhanced accuracy.

Furthermore, our study focuses on formation energy; extension to other critical materials properties such as band gaps, elastic moduli, and magnetism remains an important future task. Integrating AL with advanced equivariant neural network architectures offers exciting opportunities to reduce data requirements.


\section{Conclusion}

We have presented a thorough, multi-system benchmark evaluating active learning strategies for training machine-learned interatomic potentials on formation energy prediction. Our study spans chemically diverse systems (carbon, silicon, iron, titanium-oxide), leverages data from two major materials databases (Materials Project, OQMD), and employs rigorous statistical analysis with multiple replicates.

Key findings include:

\begin{itemize}
    \item Diversity sampling consistently yields the strongest or competitive performance, delivering particularly large gains for complex systems (e.g., a 10.9\% MAE improvement for Ti--O, $p = 0.008$).
    
    \item Uncertainty sampling provides moderate benefits, but its effectiveness is system-dependent, influenced by material class and structural complexity.
    
    \item The entire process is computationally accessible, running on commodity resources (Google Colab, less than 8 GB RAM, under 4 hours), broadening participation in materials machine learning research.
\end{itemize}

Our open-source implementation and detailed experimental protocols provide a robust foundation for community validation and further extension. We anticipate that strategically combining active learning with advanced representations will establish new milestones in data-efficient, scalable MLIP training and accelerate computational materials discovery.

\begin{center}
\rule{0.8\textwidth}{0.5pt}
\end{center}


\end{document}